\documentclass[10pt,twocolumn,letterpaper]{article}

\usepackage{cvpr}              

\usepackage{pifont}
\usepackage{times}
\usepackage{epsfig}
\usepackage{graphicx}
\usepackage{amsmath}
\usepackage{amssymb}
\usepackage{diagbox}
\usepackage{makecell}
\usepackage{booktabs}
\usepackage{indentfirst} 
\usepackage{amsmath, amssymb}
\usepackage{multirow}
\usepackage{bm}
\usepackage{mathtools}
\usepackage[pagebackref,breaklinks,colorlinks]{hyperref}

\usepackage[capitalize]{cleveref}
\crefname{section}{Sec.}{Secs.}
\Crefname{section}{Section}{Sections}
\Crefname{table}{Table}{Tables}
\crefname{table}{Tab.}{Tabs.}

\def\cvprPaperID{1848} 
\def\confName{CVPR}
\def\confYear{2022}

\begin{document}

\title{SuperGF: Unifying Local and Global Features for Visual Localization}
\author{Wenzheng Song$^{1,2}$ ~~~~Ran Yan$^{1}$ ~~~~Boshu Lei$^{1,4}$~~~~Takayuki Okatani$^{2,3}$ \\
$^{1}$Megvii ~~~~$^{2}$GSIS, Tohoku University ~~~~$^{3}$RIKEN Center for AIP ~~~~$^{4}$Xi'an Jiaotong University\\
{\tt\small \{song, okatani\}@vision.is.tohoku.ac.jp}~~~~
{\tt\small {yanran@megvii.com}}~~~~
{\tt\small {sobremesa121@gmail.com}}
}

\maketitle

\begin{abstract}
Advanced visual localization techniques encompass image retrieval challenges and 6 Degree-of-Freedom (DoF) camera pose estimation, such as hierarchical localization. Thus, they must extract global and local features from input images. Previous methods have achieved this through resource-intensive or accuracy-reducing means, such as combinatorial pipelines or multi-task distillation. In this study, we present a novel method called SuperGF, which effectively unifies local and global features for visual localization, leading to a higher trade-off between localization accuracy and computational efficiency. Specifically, SuperGF is a transformer-based aggregation model that operates directly on image-matching-specific local features and generates global features for retrieval. We conduct experimental evaluations of our method in terms of both accuracy and efficiency, demonstrating its advantages over other methods. We also provide implementations of SuperGF using various types of local features, including dense and sparse learning-based or hand-crafted descriptors.

\end{abstract}

\section{Introduction}
\label{sec:intro}
Visual localization is a key component in computer vision tasks such as Structure-from-Motion (SfM) or SLAM, which is a fundamental problem in numerous applications, such as autonomous driving, mobile robotics, or augmented reality. This growing range of applications of visual localization calls for reliable operation in both large-scale changing indoor and outdoor environments, irrespective of the weather, illumination, or seasonal changes. Visual localization is the problem of estimating the 6 Degree-of-Freedom (DoF) camera pose from which a given image was taken relative to a reference scene representation.

Advanced visual localization approaches are hierarchical, encapsulating image retrieval problems and 6-DoF camera pose estimation~\cite{hf-net, yang2018deep, taira2018inloc}. i.e., given a query image of the current view, related candidates are determined by performing image retrieval in a database; then, perform local feature matching between the query image and related candidates to establish point correspondences between 2D to 3D, solving a PnP and estimate 6 DoF camera pose of the query image.
\begin{figure}[t]
\centering
\includegraphics[width=0.95\linewidth]{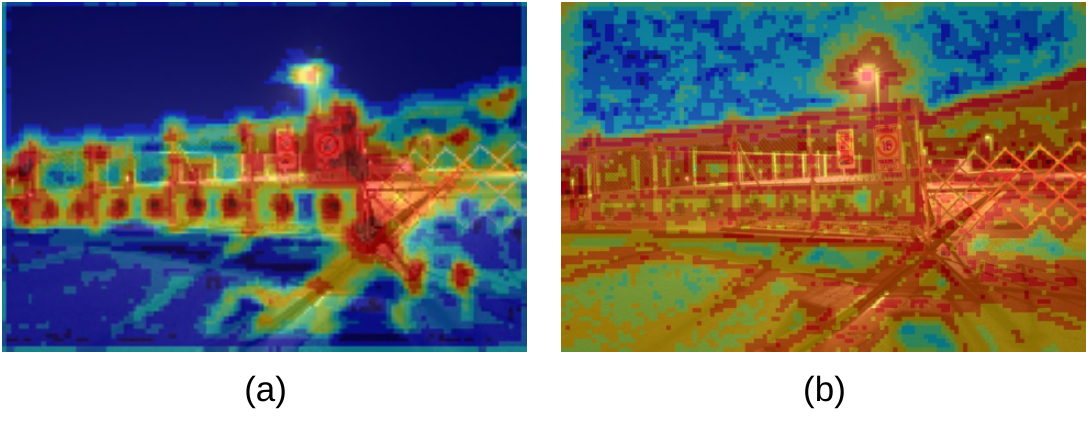}
   \caption{Illustration of the feature-level gap between the two tasks. i.e., (a) image retrieval and (b) image matching. Where (a) and (b) are activation maps of local features generated by NetVLAD~\cite{arandjelovic2016netvlad} and SuperPoint~\cite{superpoint}, respectively.}
\label{fig:feature_gap}
\end{figure}
Naturally, two types of image features are needed to perform such a hierarchical visual localization approach, i.e., global image features for retrieval and local image features for image-matching. However, existing studies struggle to unify these two types of image features. For instance, the image retrieval task also involves these two types of features for the two-stage retrieval strategy~\cite{cao2020unifying}, i.e., global features are used for large-scale coarse retrieval to get prior candidates, then re-rank it based on the number of inliers obtained from local feature matching with geometry verification. However, such local features tend to deviate from needs from 6-DoF pose estimation. Specifically, local features yielded by a retrieval model always tend to be semantically rich and lose spatial invariance~\cite{ng2020solar,ge2020self}. Moreover, retrieval models don’t provide key-point detection, so these local features lack corresponding exact pixel coordinate information. 

Most studies employ two models for the hierarchical visual localization., i.e., one for image retrieval, such as NetVLAD~\cite{arandjelovic2016netvlad}, and another for image matching, such as SIFT~\cite{sift} or SuperPoint~\cite{superpoint}. However, such a method is not competitive as far as efficiency is concerned due to the absence of shared computing. In addition, it is difficult to trade off efficiency and accuracy in model selection. Specifically, accurate image retrieval can improve the accuracy and efficiency of localization within a certain range. However, advanced image retrieval models tend to be heavy to pursue higher accuracy, which causes an increase in computational cost. Still, these improvements do not result in corresponding gains for localization.

A recent study unifies global and local features for visual localization using multitask distillation, i.e., HF-Net~\cite{hf-net}. It uses a CNN that jointly estimates local and global features with a shared encoder. It improves computational efficiency by using a compression model, i.e., MobileNetVLAD (MNV). However, the distillation inevitably harms inference accuracy, especially in challenging cases, e.g., night scenes. One possible reason is that they do not consider the conflict between the two tasks, i.e., image retrieval and image-matching, at the feature level; see Figure~\ref{fig:feature_gap}. Specifically, learning global features for image retrieval tends to induce features in the backbone to be more semantic and less localizable, eventually making it more sparse. Conversely, learning local features for image-matching emphasizes feature invariance, which leads the features to focus more on local information and lacks global semantics. This issue has been demonstrated earlier~\cite{cao2020unifying} but needs to be handled more appropriately.

Most recently, transformers have been used for feature extraction in computer vision tasks and have led to state-of-the-art results~\cite{ng2020solar,wang2022transvpr}. It benefits from the desirable property of the self-attention mechanism, which can break through spatial constraints to establish global correlations, and aggregate task-relevant features naturally. 

In this paper, we propose a novel holistic model, SuperGF, which unifies global and local features for visual localization. It works directly on the local features generated by the image-matching model and aggregates to a global image feature, similar to BoW~\cite{BoW,ret2} or Fisher Vector~\cite{fisher}. A transformer is adopted to perform feature aggregation, which is more accurate and resource-friendly. It integrates global contextual information and establishes global correlations between feature tokens by the self-attention mechanism; thus, semantic cues can be learned automatically from local features of the image-matching model. As a result, the transformer module can bridge the feature-level gap between the two tasks and yield robust global features for image retrieval in an efficient manner. We experimentally evaluate global features yield by SuperGF on several benchmarks. Then we also assess the performance of visual localization using the holistic model combining different kinds of local features. The results show the advantage of our model compared to existing methods.

\section{Related Work}
\label{sec:relat}
\subsection{Approaches for Visual Localization}
\label{sec:relat_l}
The approaches of previous works on visual localization can be summarized as follow:
\vskip.5\baselineskip

\noindent \textbf{Structure-based Localization.} Previous visual localization approaches mainly rely on estimating correspondences between 2D keypoints in the query and 3D points in a sparse model using local descriptors. The map is usually composed of a 3D point cloud constructed via Structure-from-Motion (SfM), where each 3D point is associated with one or more local feature descriptors. The query pose is obtained by feature matching and solving a Perspective-n-Point problem (PnP)~\cite{lepetit2009epnp}. However, direct matching methods tend to be resource-intensive or fragile and challenging to apply in large-scale localization.
\vskip.5\baselineskip

\noindent \textbf{Image-based Localization.} Visual localization in large-scale urban environments is often approached as an image retrieval problem. Specifically, the location of a given query image is predicted by transferring the geotag of the most similar image retrieved from a geotagged database~\cite{arandjelovic2016netvlad,arandjelovic2014dislocation,chen2011residual,jin2017learned,sattler2016large,torii201524,torii2013visual}. This approach scales to entire cities
thanks to compact image descriptors and efficient indexing techniques~\cite{RootSIFT,superglue,d2net,r2d2} and can be further improved by spatial re-ranking~\cite{cao2020unifying}, informative feature selection~\cite{chum2007total,chum2011total} or feature weighting~\cite{gronat2013learning,jegou2009burstiness,sattler2016large,torii2013visual}. Image-based localization approaches have recently shown promising results in robustness and efficiency but are not competitive in terms of accuracy~\cite{arandjelovic2016netvlad,gcl}, which output only an approximate location of the query, not an exact 6-DoF pose.
\vskip.5\baselineskip

\noindent \textbf{Hierarchical Localization.} Hierarchical localization takes an approach, dividing the problem into a global, coarse search followed by a fine pose estimation. It shows advantages in terms of efficiency and accuracy compared to the above two approaches, which can be applied to large-scale. The intermediate retrieval step of hierarchical localization limits the downstream feature matching to a reasonable range, which reduces the computational cost significantly while improving the localization performance by reducing the influence of feature repetition.  \cite{sarlin2018leveraging} proposed to search at the map level using image retrieval and localize by matching hand-crafted local features against retrieved 3D points. However, its robustness and efficiency are limited by the underlying local descriptors and heterogeneous structure. Taira~\etal applied learning-based features to camera pose estimation but in a dense, expensive manner~\cite{taira2018inloc}. Recently,  HF-Net~\cite{hf-net} integrated learning-based models of image retrieval and image-matching, by model distillation that simultaneously predicts keypoints as well as global and local descriptors for accurate 6-DoF localization.

\subsection{Global and Local Image Features }
Before the emergence of deep learning, hand-crafted local features, such as SIFT~\cite{sift}, ORB~\cite{orb}, and SURF~\cite{surf}, are widely applied in computer vision fields such as image matching. Moreover, traditional aggregation methods~\cite{ret2, fisher} are developed for generating global image features for image retrieval using these hand-crafted local features. However, hand-crafted local features are limited in invariance due to only involving low-level information. 

Recent features emerging from convolutional neural networks (CNN) exhibit unrivaled robustness at a low computing cost. However, it tends to be task-specific. Specifically, task-specific local or global image features are generated using different models in an end-to-end manner. Even though they achieve superior performances in their respective domain, such as image retrieval~\cite{ng2020solar,wang2022transvpr,hausler2021patch, cao2020unifying} or image matching~\cite{superpoint,superglue,d2net,r2d2,lfnet}, there are still problems in unifying for multi-task.

More recently, transformers have been adopted for feature extraction in computer vision fields and achieved state-of-art performances~\cite{ng2020solar, wang2022transvpr}. It benefits from the desirable property of the self-attention mechanism, which can naturally aggregate task-relevant features. Recent studies have applied transformers to each component method for visual localization, i.e., image retrieval~\cite{ng2020solar, wang2022transvpr,superfeature} and image matching~\cite{superglue,sun2021loftr}.

\section{Unify Local and Global features for Visual Localization.}
\label{sec:method}
\subsection{Overview}
\begin{figure*}[t]
\centering
\includegraphics[width=0.95\textwidth]{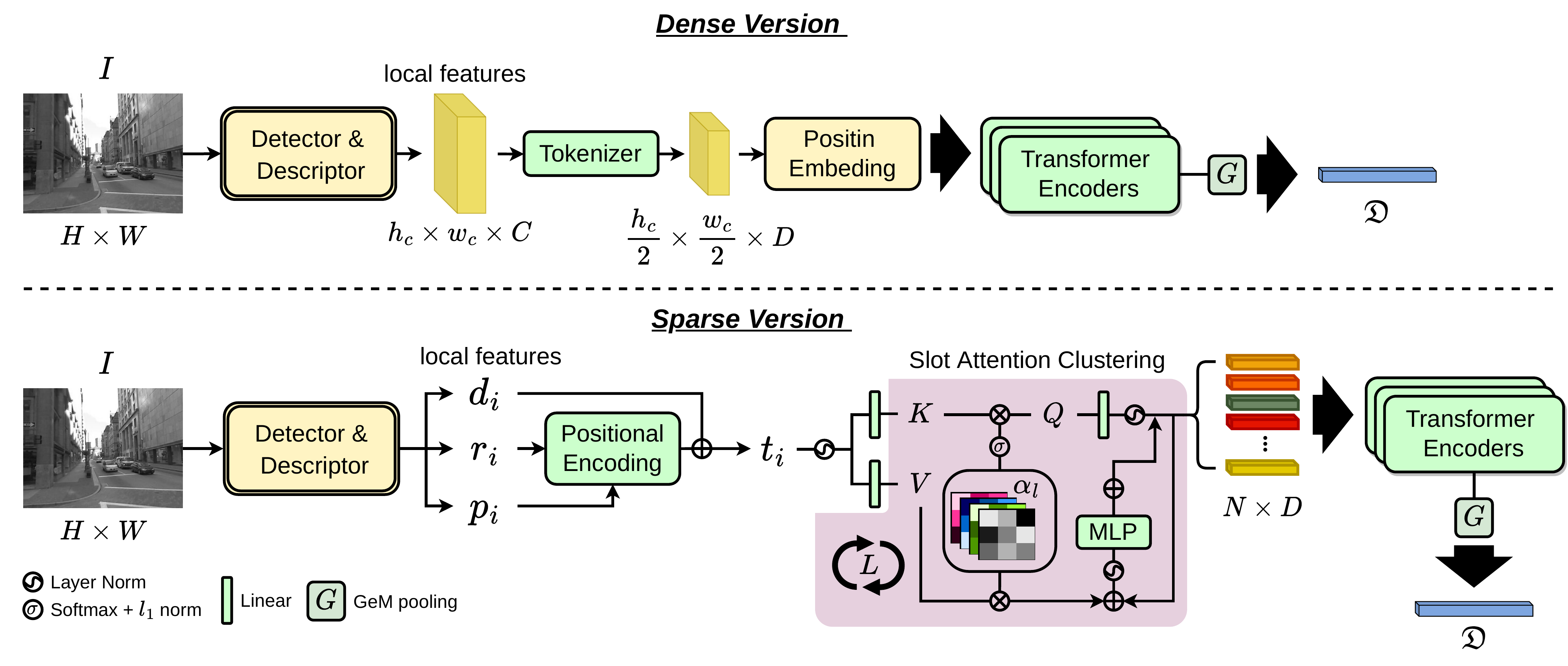}
   \caption{The framework of SuperGF. We provide two implementations of SuperGF, i.e., the dense version and the sparse version. The former works on dense local descriptors, such as dense SIFT or feature maps output by the SuperPoint encoder. The latter works on sparse local features used for image matching, i.e., keypoints ($p_i$), descriptors ($d_i$), and confidence scores ($r_i$). The modules indicated by {\color{green}green baskets} contain learnable parameters.}
\label{fig:pipeline}
\end{figure*}

SuperGF essentially plays the role of feature aggregation, i.e., aggregate local image features used for image-matching, which are more localizable but lack semantics, into glocal image features with rich semantic information used for image retrieval. Figure.~\ref{fig:pipeline} illustrates the framework of SuperGF.

Given an input image, local features are extracted by a detector and descriptor at first. In practice,  SuperGF is designed for both a hand-crafted descriptor and a learning-based descriptor, i.e., SIFT~\cite{sift} and SuperPoint~\cite{superpoint}, respectively. Moreover, we consider two different forms of the descriptors separately, i.e., dense or sparse. Specifically, in the case of the sparse version, only local descriptors where keypoints are taken into count. In the case of the dense version, we adopt dense descriptors, i.e., dense SIFT or feature map output by the encoder of SuperPoint.

Then, we process the input of local features into tokens. In the case of the dense version, we perform the local descriptors by a tokenizer consisting of a one-layer CNN with downsampling operations. Then, we perform position embedding for these tokens.  In the case of the sparse version, the input of local features is a variable-length sequence. To save computational costs, we downsample them by performing clustering. Before that, we integrate three features of descriptors, keypoints, and confidence scores by performing positional encoding.

Finally, these tokens generated by the previous step are input into the three layers of the vision transformer encoder. And a global image feature is aggregated from the transformer's output using a GeM pooling layer. 

\subsection{Local Feature Processing}
\label{sec:Mulatt}
\noindent \textbf{Dense version.} Given an input image $I\in\mathbb{R}^{H\times{W}}$, the feature map of local descriptors of $I$ can be denoted as $\mathbf{F}\in\mathbb{R}^{{128}\times\frac{H}{8}\times\frac{W}{8}}$. Let's define $\mathcal{F}: \xi \rightarrow \mathbb{R}^{D\times\frac{H}{16}\times\frac{W}{16}}$ to denote a CNN layer encodes feature map $\mathbf{F} \in \xi$ into a feature with the shape of $(D, \frac{H}{16}, \frac{W}{16})$, then, the output of the tokenizer can be denoted as:
\begin{equation}
\bm{T}_{raw} = ReLU(\mathcal{F}(\mathbf{F}))+ \bm{\mathcal{P}}
\end{equation}
where $\bm{T}_{raw},\bm{\mathcal{P}}\in\mathbb{R}^{D\times\frac{H}{16}\times\frac{W}{16}}$, $\bm{\mathcal{P}}$ is the tokens of \textit{Position Embedding}, $D$ means the number of channels of raw patch tokens, specifically, $D=512$. Finally, the raw patch tokens $\bm{T}_{raw}$ is reshaped into a sequence of flattened 2D pattern $(\frac{H}{16}\times\frac{W}{16}, D)$ as input tokens for transformer encoder layers. In practice, $7\times7$ convolutional kernel is used, and the stride is $2$.
\vskip.5\baselineskip

\noindent \textbf{Sparse version.} Given an input image $I\in\mathbb{R}^{H\times{W}}$, the sparse local features of $I$ can be denoted as $d_i\in\mathbb{R}^{i\times d}$: local descriptors, $p_i\in\mathbb{R}^{i\times 2}$: keypoints position, and $r_i\in\mathbb{R}^{i\times 1}$: confidence scores. $d=128$ when using SIFT, and $d=256$ when using SuperPoint. 

The initial representation $t_i\in\mathbb{R}^{i\times d}$ for each keypoint $i$ combines its visual appearance and location. We embed the keypoint position into a high-dimensional vector with a Multilayer Perceptron (MLP) as:
\begin{equation}
t_i = d_i + MLP_{enc}(p_i, r_i)
\end{equation}
This encoder enables the graph network to later reason about both appearance and position jointly, especially when combined with attention, called \textit{Positional Encoding}.

Inspired by recent studies~\cite{}, we propose a clustering module base on slot attention, which takes an input of $t_i$ and outputs an ordered set of features of clusters, denoted by $\boldsymbol{\zeta}$. Let the clustering module be represented by function $\Phi(\bullet): \mathbb{R}^{i\times d} \to \mathbb{R}^{N\times D}$, which can be defined as an \textit{iterative} module:
\begin{equation}
\Phi(\bullet) = \boldsymbol{\zeta}^L, \quad \boldsymbol{\zeta}^l = \phi(\bullet;\boldsymbol{\zeta}^{l-1})
\end{equation}
where $\phi$ denotes the core function of the module applied $L$ times, and $\boldsymbol{\zeta}^0 \in \mathbb{R}^{N\times D}$ denotes a set of learnable templates of clusters, which are initialized randomly. In practice, we set $N=512$ and $L=6$. The final output is progressively formed by iterative refinement of the templates and each token of $\boldsymbol{\zeta}$ is a function of all input of $t_i$.

The architecture of the core function $\phi$ is inspired by~\cite{vaswani2017attention,vit} and is composed of a dot-product attention function $\psi$, followed by an MLP. The function $\psi$ receives three inputs, i.e., \textit{key}, \textit{value}, and \textit{query}, represented by $K$, $Q$, and $V$, respectively. The corresponding input generates the $K$, $Q$, and $V$, passing through layer normalization and fed to three linear projection functions that project them to dimensions $d_k$, $d_q$, and $d_v$, respectively. In practice, we set $d_k =d_q=d_v=D=512$, $K$ and $V$ are generated by $t_i$, and Q is generated by $\boldsymbol{\zeta}$. The functions $\phi$ and $\psi$ is given by:
\begin{equation}
\phi(\bullet) = \psi(\bullet)+MLP(\psi(\bullet)), \quad \psi(\bullet)= \boldsymbol{\alpha}\cdot V +\boldsymbol{\zeta}^l
\end{equation}
where $\boldsymbol{\alpha}$ denotes the attention maps over $t_i$. Thus, we have $N$ attention maps in total. i.e., $\boldsymbol{\alpha}\in \mathbb{R}^{i\times N}$ $\boldsymbol{\alpha}$ can be calculated as follow:
\begin{equation}
\boldsymbol{\alpha} = \mathcal{N}_{L_1}(\mathrm{softmax}(\frac{QK^\top}{\sqrt{d_k}}))
\end{equation}
where $\mathcal{N}_{L_1}$ denotes $L_1$ normalization. 

\subsection{Aggregate to Global Image Features}

SuperGF employs vision transformer encoders to achieve feature aggregation, integrating global contextual information and establishing global correlations between feature tokens by self-attention mechanism, and semantic cues can be learned automatically from the input of local features. In practice, we follow the implementation of the vision transformer encoder, which consists of a stack of Multi-headed Self-Attention (MSA) and Multi-Layer Perceptron (MLP) modules \cite{vaswani2017attention,vit}, and add skip-connection between two transformer encoders. Moreover, we discarded the operation of \textit{patch embedding} and instead it by the operation of the previous step; see Sec.~\ref{sec:Mulatt}. Let’s define the input as $\mathbf{T}_0$, where $\mathbf{T}_0$ denotes the output of the previous step, i.e., $\mathbf{T}_{raw}$ or $\boldsymbol{\zeta}^{L}$. Then define function $\bm{\Theta}_n(\bullet): n \in [1,4]$ to denote the $n$-layer of vision transformer encoders. The output tokens $\mathbf{T}_{out}$ is given by:
\begin{equation}
\begin{split}
    &\bm{T}_n =\sum_{n=1}^3 \left[ \bm{T}_{n-1} + \bm{\Theta}_n(\bm{T}_{n-1})\right],\\
    &\mathbf{T}_{out} = MLP(\mathcal{L}(\bm{T}_3))
\end{split}
\end{equation}
where $\mathcal{L}$ denotes the layer normalization, In practice, we double the dimension of the input in the MLP. Thus, $\mathbf{T}_{out} \in \mathbb{R}^{(\frac{H}{16}\times\frac{W}{16})\times 1024}$ in the case of the dense version or $\mathbf{T}_{out} \in \mathbb{R}^{N\times 1024}$ in the case of the sparse version. As a result, $\mathbf{T}_{out}$ tends to focus on task-relevant local features.

Finally, $\mathbf{T}_{out}$ is aggregated to global image features by performing  Generalized Mean (GeM) pooling ~\cite{}, which is a flexible way to aggregate local features with a learnable parameter $p$. The aggregated global image feature $\boldsymbol{\mathfrak{D}}$ is given by:
\begin{equation}
\begin{split}
&\boldsymbol{\mathfrak{D}} = \mathcal{N}_{L_2}(GeM(\mathbf{T}_{out},p)), \\
&GeM(\mathbf{T}_{out},p) = \left[\frac{1}{\lambda}\sum_{i=0}^\lambda{(T_{out}^i)^p}\right]^{\frac{1}{p}}
\end{split}
\end{equation}
where $\boldsymbol{\mathfrak{D}} \in \mathbb{R}^{1 \times 1024}$, $\mathcal{N}_{L_2}$ denotes $L_2$ normalization, and $\lambda$ denotes the special size of $\mathbf{T}_{out}$, i.e., $\lambda=\frac{H}{16}\times\frac{W}{16}$ in the case of the dense version and $\lambda=N$ in the case of the sparse version. In practice, we set $p=3$ in parameter initialization.
\subsection{Training Strategy}
\subsubsection{Loss Function}
In most studies, image retrieval or Visual Place Recognition (VPR) is treated as a binary problem. Models are trained using pairs or triplets of image samples labeled as either positive or negative with a metric learning loss, such as the triplet loss~\cite{triplet,dong2018triplet} or a contrastive loss~\cite{wang2021understanding}. However as mentioned in existing studies~\cite{gcl, ap_re}, since the distribution of scenes in the real world is continuous, in the VPR task, the similarity between images in VPR cannot be strictly defined in a binary fashion. Thus as we observed, in most VPR datasets, there is a large ambiguity between the positive and negative samples of the training data.

Moreover, VPR is essentially a problem of ranking, not classification. For training a VPR model, metric learning losses distinguish between positive and negative samples by defining a margin, which easily leads to a distance ambiguity. This results in the performance being very sensitive to the loss settings so it is often difficult to achieve optimal performance by optimizing a metric learning loss directly. 

\begin{figure}[t]
\centering
\includegraphics[width=0.95\linewidth]{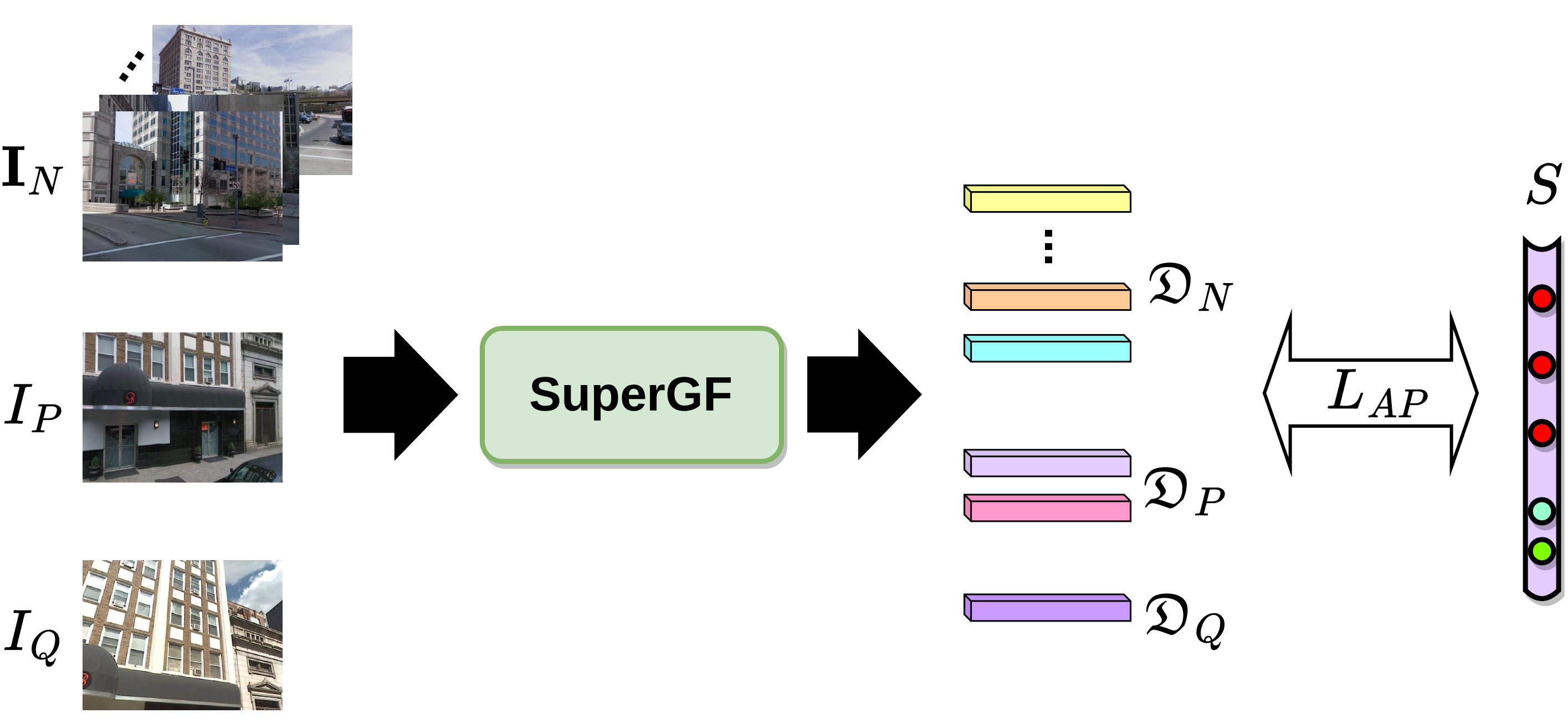}
   \caption{ Illustration of the training strategy. Each sample of training data includes a batch composed of one query image $I_Q$, one positive sample $I_P$, and $\alpha+\beta$ negatives; of these, $\alpha$ soft negative samples and $\beta$ negative samples. i.e., $I_N^i \in \bm{I}_N : {i} = [1,...,\alpha + \beta]$. In practice, $\alpha=2$ and $\beta=100$.}
   
\label{fig:train}
\end{figure}

Considering the above, we adopt a listwise loss. i.e., Average Precision (AP) loss~\cite{chen2020ap,ap_re,r2d2}, to train our model. The AP loss train models by optimizing ranking results directly. We follow the implementation of AP loss as Chen~\etal~\cite{ap_re}. As illustrated in Fig.~\ref{fig:train}, after inference an image batch of $[I_Q, I_P, \bm{I}_N]$, we obtain global image descriptors $[\boldsymbol{\mathfrak{D}}_Q, \boldsymbol{\mathfrak{D}}_P, \boldsymbol{\mathfrak{D}}_N]$. Then we calculate cosine similarities $\bm{\mathcal{S}} = \mathcal{S}_i: {i}\in [1,\alpha+\beta]$ between $\boldsymbol{\mathfrak{D}}_Q$ and $\boldsymbol{\mathfrak{D}}_P\cup{\boldsymbol{\mathfrak{D}}_N}$: 
\begin{equation}
\mathcal{S}_i = sim(\boldsymbol{\mathfrak{D}}_Q, \boldsymbol{\mathfrak{D}}^i) = \boldsymbol{\mathfrak{D}}_Q^\top  \boldsymbol{\mathfrak{D}}^i
\end{equation}
where $\boldsymbol{\mathfrak{D}}^i \in \boldsymbol{\mathfrak{D}}_P\cup{\boldsymbol{\mathfrak{D}}_N}$, $\mathcal{S}_i \in [-1, 1]$. Let's define the function $AP(\cdot)$ denotes the ranking metric AP for $\boldsymbol{\mathfrak{D}}_Q$, The final loss function can be defined as:
\begin{equation}
    L_{AP} = 1-AP(\bm{\mathcal{S}},\bm{S}) 
\label{eq:6}
\end{equation}
where $\bm{S} \in [0, 1]$ denotes the label of image similarity scores between $I_Q$ and $I_P \cup{\bm{I}_N}$. 

In the case of the sparse version, we further adopt an attention decorrelation loss which aims at reducing the spatial correlation between attention maps. It makes the output of the slot attention clustering module, i.e., $\boldsymbol{\zeta}$, as complementary as possible. Specifically, we encourage them to attend to different local features, i.e., different locations of the image. Let's define $\boldsymbol{\alpha}= [\alpha_1,...,\alpha_N]$, The attention decorrelation loss is given by: 
\begin{equation}
        L_{attn} = \frac{1}{N(N-1)}\sum_{i\neq j}\frac{\alpha_i\cdot\alpha_j}{\left \| \alpha_i \right \|_2\left \| \alpha_j \right \|_2}
\end{equation}
where $i,j\in \left\{1,...,N\right\}$. In practice, We train our sparse version model with the above two losses jointly.

\subsubsection{Training Setting}
\label{sec:Training Setting}
First, we aim to design robust global image descriptors that can be used for place recognition for SuperPiont dependencies. So we fixed the SuperPoint encoder during the training process. We perform image-level supervision using the AP loss. In practice, we adopt the data annotation provided by Vallina \etal. \cite{gcl} as the soft label, i.e, $\bm{S}$ for our model training, which represents Field-of-View (FoV) overlap between query and database images. 

Specifically, for each query image, we split the database images into three kinds based on FoV overlap, i.e., positive samples where $S\in[0.5, 1]$, soft negative samples where $S\in(0, 0.5)$, and negative samples where $S=0$. For each query image $I_Q$, we compute the AP loss between it with 105 images in the database, which contains 1 positive sample, 2 soft negative samples and 100 negative samples. For each iteration of the training process, we load 4 batches to compute loss and optimize the parameters of our model. We observe that choosing more negative samples for each query image tends to result in better performance. However, selecting too many negative samples takes up more computational resources and provides limited gain to the model.

\section{Experiments}
We experimentally evaluate the proposed SuperGF on several benchmark datasets compared with some state-of-the-art methods. We evaluate our method on two tasks, i.e., VPR and visual localization. The former aims mainly to examine the performance of global image features generated by SuperGF. We give the details of experimental settings, datasets, evaluation metrics, and compared methods in the following.

\subsection{Implementation details}
\label{sec:setting}
\noindent\textbf{Model Settings.}  We set the latent embedding dimension of the transformer encoders as 512, and the hidden dimension of MLP blocks as 1024. We trained our model on the MSLS training set with settings reported in Sec.~\ref{sec:Training Setting}. We selected 10,000 query samples randomly from all sub-cities for one epoch and trained our model on 350 epochs in total. We adopted the AdamW optimizer and set the parameter of weight\underline{\hspace{0.5em}}decay $=10^{-4}$ in training. We set an initial learning rate to $10^{-4}$ and it will finally decline to $10^{-6}$. We set the input image size to $640\times 480$.
\vskip.5\baselineskip

\noindent\textbf{Experiment Settings for VPR.} We perform the retrieval based on the $L_2$ distance between global image features to obtain the prior candidates. We further perform geometry verification by matching sparse descriptors provided by the front-end model, i.e., SIFT or SuperPoint. In practice, given an image pair of the query and database, we obtain keypoints and sparse local descriptors for both. Then, we perform feature matching to obtain initial point correspondences, using Nearest Neighbor searching (NN) or SuperGlue~\cite{superglue}. Finally, we estimate the homography with RANSAC as a robust estimator and re-rank the prior candidates based on the number of inliers. We keep the top 100 candidates retrieved by global image features and perform the geometry verification.
\vskip.5\baselineskip

\noindent\textbf{Experiment Settings for Localization.}
Firstly, a sparse reference model is built with SfM using database images. Specifically, we adopt prebuilt reference models with COLMAP using RootSIFT or SuperPoint, provided by~\cite{robotcar} and~\cite{hf-net}. They are each used in the case of the corresponding descriptor employed in SuperGF to avoid the influence of using different detectors on feature matching. For localization, given a query image, we retrieve the top-$n$ candidates from the database using global image features generated by SuperGF and match them one by one. In practice, we adopt NN and also SuperGlue when SuperPoint is selected. Then, we keep all matches where a corresponding 3D point is included in the reference SfM model. Finally, we can lift them to 2D-3D matches, and the 6-DoF camera pose of the query image is obtained by solving PnP with RANSAC.





\subsection{Evaluation Datasets}
\label{Evaluation Datasets}
We evaluated our method on several public benchmark datasets: MSLS\cite{warburg2020mapillary}, Pitts30k \cite{torii2013visual}, Nordland \cite{olid2018single} and Tokyo247 \cite{torii201524} for VPR; The RobotCar Seasons dataset \cite{robotcar} for localization. All of these datasets contain different challenging appearance variations, such as changes in day-night, weather, and season. More details of dataset usage are given in Supplementary Material. All images are resized to $640\times 480$ while evaluation.

\subsection{Metrics}
As illustrated in Sec.~\ref{Evaluation Datasets}, for VPR datasets, we use Recall@$N$ metric, which computes the percentage of query images that are correctly localized. A query image is considered retrieved successfully if at least one of the top $N$ ranked reference images is within a threshold distance from the ground truth location of the query image. Default threshold definitions are used for all datasets \cite{hausler2021patch,wang2022transvpr,arandjelovic2016netvlad}. 

For the localization dataset, i.e., the RobotCar Seasons dataset, we report the pose recall at position and orientation thresholds different for each sequence, as defined by the benchmark~\cite{robotcar}. 

\subsection{Compared Methods}
We experimentally compared SuperGF against several models on both VPR and visual localization tasks with settings introduced in Sec.~\ref{sec:setting}. 
\vskip.5\baselineskip
\noindent\textbf{VPR methods.} We compare methods in the VPR task considering both two approaches, i.e., single-pass and two-stage retrieval. The former only uses global image features, and the latter performs re-ranking with geometry verification additionally. For the experiment of single-pass retrieval, we compare with a widely used baseline, i.e., NetVLAD~\cite{arandjelovic2016netvlad}, several state-of-the-art CNN-based methods, including SFRS~\cite{ge2020self}, ResNet50-GeM-GCL~\cite{gcl}, GeM-AP~\cite{ap_re}, and also two state-of-the-art transformer-based methods, i.e., SOLAR~\cite{ng2020solar} and TransVPR~\cite{wang2022transvpr}. For the experiment of two-stage retrieval, we compare with DELG~\cite{cao2020unifying}, and two state-of-the-art methods, i.e., Patch-NetVLAD~\cite{hausler2021patch} and TransVPR. As mentioned earlier in the paper, local features generated by all the above methods don’t support 6-DoF camera pose estimation. In addition, we further compare with a strong hybrid baseline, NV-SP-SG, which re-ranks NetVLAD retrieved candidates by using SuperGlue matcher to match SuperPoint local features.

\vskip.5\baselineskip
\noindent\textbf{Localization methods.} 
Since image-based localization approaches lack accuracy, we only consider localization approaches that can obtain 6-DoF camera poses; see Sec.~\ref{sec:relat_l}. In practice, we compare ours with two methods based on structure-based localization approaches, i.e., Active Search (AS)~\cite{sattler2016efficient} and City Scale Localization (CSL)~\cite{svarm2016city}, hierarchical localization approach based on hybrid pipelines including NV+SIFT, NV+SP, and also HF-Net~\cite{hf-net}. NV+SIFT and NV+SP mean using global features extracted by NetVLAD and local features extracted by SIFT or SuperPoint. 

\section{Results and Discussion}
\subsection{Single-pass Retrieval}
Table~\ref{tab:VPR_results} shows the experimental results for single-pass retrieval. We can make the following observations. 
\begin{table*}\centering
    \caption{Single-pass retrieval results with three metrics of recall, where 'R-' means 'Recall'.} %
    \label{tab:VPR_results}
    \resizebox{1.0\textwidth}{!}{
    \begin{tabular}{*{1}{l}||*{3}{c}||*{3}{c}||*{3}{c}||*{3}{c}||*{3}{c}}
        \toprule
        \multicolumn{1}{c||}{}& \multicolumn{3}{c||}{MSLS val}&\multicolumn{3}{c||}{MSLS challenge}&\multicolumn{3}{c||}{Pitts30k test}&\multicolumn{3}{c||}{Nordland test}&\multicolumn{3}{c}{Tokyo247 test}\\\cline{2-16}
        Method & R@1 &R@5& R@10  & R@1 &R@5& R@10   & R@1 &R@5& R@10&  R@1 &R@5& R@10&  R@1 &R@5& R@10\\\midrule
        NetVLAD & 47.7 & 62.8& 70.9 & 30.7& 41.9 &46.4 &  82.1 & 91.4 & 93.8  & 12.5& 21.4&26.1  & 57.1 & 72.1 & 77.1\\
         SFRS &  69.2 &80.3& 83.1   &  41.5 &52.0 &56.3    &  \textbf{89.4} &\textbf{94.7} &\textbf{95.9} &  18.8& 32.8 &39.8 &   \textbf{85.4}& \textbf{91.1}& \textbf{93.3}\\
          ResNet50-GeM-GCL & 66.2	&78.9	&81.9  & 43.3&	59.1&	65.0  & 72.3&	87.2&	91.3 & 27.2 & 41.1 &49.2 &44.1&	61.0&	66.7\\
           GeM-AP &  64.1& 	75.0&	78.2  &  33.7&	44.5&	49.4   &  80.7	&91.4&	94.0 & 11.8 & 18.4 & 22.7 &  11.4&	22.9&	30.5\\
        SOLAR & \textbf{78.3} &\textbf{87.2}&\textbf{89.6} & 45.2 &60.5& 69.3  & 85.4 &92.6& 94.8 &36.6  &51.3 & 58.8& 76.2  &84.4&88.3\\
        TranVPR &  70.8 &85.1 &\textbf{89.6}  &  48.0 &\textbf{67.1}&\textbf{73.6}    &  73.8 &88.1& 91.9 & 15.9 &38.6 &49.4 & --  &--&--\\\hline
        SuperGF-DenseSP & 72.8 & 81.5& 85.2  & \textbf{50.8} &66.0&  71.6  & 68.4 & 82.5& 87.3& \textbf{58.4} &\textbf{78.6}& \textbf{85.5}& 23.2 &37.8&46.0\\
        SuperGF-SparseSP & 71.3 & 81.3 & 85.0  & 48.2 & 65.3& 70.4& 60.2 &77.6& 83.1    &57.1  &75.4&82.0 &  25.7  &39.4&44.4\\
        SuperGF-DenseSIFT & 60.2 & 72.8 & 75.3  & 39.7&56.2 & 63.9& 53.2 &71.6& 78.1    &35.6  &50.7&56.7 &  14.7  &26.8&36.3\\
        SuperGF-SparseSIFT & 49.7 & 67.5 & 74.4  &11.2 & 22.7& 37.5& 37.7 &59.1& 69.2    &15.5  &30.5&38.4 &  8.6 &15.9&21.0
\\\bottomrule
\end{tabular}}
\end{table*}

\begin{table*}\centering
    \caption{Two-stage retrieval results with three metrics of recall, where 'R-' means 'Recall'.} %
    \label{tab:VPR_two}
    \resizebox{1.0\textwidth}{!}{
    \begin{tabular}{*{1}{l}||*{3}{c}||*{3}{c}||*{3}{c}||*{3}{c}||*{3}{c}}
        \toprule
        \multicolumn{1}{c||}{}& \multicolumn{3}{c||}{MSLS val}&\multicolumn{3}{c||}{MSLS challenge}&\multicolumn{3}{c||}{Pitts30k test}&\multicolumn{3}{c||}{Nordland test}&\multicolumn{3}{c}{Tokyo247 test}\\\cline{2-16}
        Method & R@1 &R@5& R@10  & R@1 &R@5& R@10   & R@1 &R@5& R@10&  R@1 &R@5& R@10&  R@1 &R@5& R@10\\\midrule
        DELG & 83.2 &90.0 &91.1  &   52.2& 61.9 &65.4  & \textbf{89.9}& \textbf{95.4} &\textbf{96.7} &  51.3 &66.8& 69.8 & 78.8&86.7&90.0\\
        NV-SP-SG & 78.1&81.9& 84.3  &  50.6 &56.9 &58.3   & 87.2 &94.8 &96.4 & 29.1 &33.5& 34.3 & 71.2 & 82.0 &78.8 \\
        Patch-NetVLAD & 79.5& 86.2 &87.7 & 48.1 &57.6 &60.5  & 88.7 &94.5& 95.9 &  46.4 &58.0& 60.4 & \textbf{95.9}&	\textbf{96.8}&	\textbf{97.1}\\
        TransVPR & 86.8 &91.2 &92.4 &63.9 &74.0 &77.5 & 89.0& 94.9& 96.2 &58.8 &75.0& 78.7& --  &--&--\\\hline
        SuperGF-DenseSP & \textbf{87.7} & \textbf{92.0} & \textbf{93.2} &\textbf{65.2} & \textbf{76.5} & \textbf{80.1} &84.6 & 92.1 &94.0& \textbf{87.4}& \textbf{95.3}& \textbf{97.2}& 68.9& 72.4 & 73.7\\
        SuperGF-SparseSP & 86.4 & 91.5 & 92.9  & 64.4& 74.3& 79.1& 83.3 &90.8& 93.8    &85.3  &94.7&97.1 &  63.2  &70.6&72.8\\
        SuperGF-DenseSIFT & 71.3 & 81.3 & 85.0  &50.2 &57.1 &59.3 & 60.2 &77.6& 83.1    &57.1  &75.4&82.0 &  25.7  &39.4&44.4\\
        SuperGF-SparseSIFT & 49.7 & 67.5 & 74.4  & 45.5 &50.2 &59.8 & 55.2 &67.1& 78.9    &46.8  &54.1&60.2 &  25.2  &37.9&43.1
\\\bottomrule
\end{tabular}
    }
\end{table*}

First, the performance of single-pass retrieval only based on global image features is highly relevant to the dataset, which denotes global features as a compact representation of the entire image, which exhibits undesirable generalizability, especially for high-precision retrievals, such as R@$1$ and R@$5$. 

Next, in general, even if we do not use separate generated task-specific local features as other methods do, i.e., we adopt local features designed for image-matching. Our method based on learned local features, such as SuperGF-denseSP or SuperGF-sparseSP, can still generate global image features on par with state-of-the-art retrieval models. It demonstrates the advantages of the transformer in feature aggregation. 

Finally, from the comparison between different implementations of our method, we can observe that using learned local features for the aggregation, i.e., SuperGF-denseSP and SuperGF-sparseSP, show significantly better results than using hand-craft local features, i.e., SuperGF-denseSIFT and SuperGF-sparseSIFT. In addition, using dense local features show better results than using sparse local features, especially for hand-craft local features.

\subsection{Two-stage Retrieval}
Table~\ref{tab:VPR_two} shows the experimental results for two-stage retrieval. We can make the following observations. 

First, compared to single-pass retrieval, the robustness is significantly improved by performing re-ranking based on geometry verification, which illustrates the importance of spatial information for image retrieval, which is discarded by global image features. 

Next, using sparse local features designed for image-matching for the re-ranking show on par even better performance than using those task-specific local features. It is noteworthy that the former performs matching sparse local features, which is more competitive in terms of efficiency than the latter, which performs matching local features densely. 

Finally, our methods with the SuperGlue matcher, i.e., SuperGF-denseSP-SG and SuperGF-sparseSP-SG, show state-of-the-art level performance compared to others.

\begin{table*}\centering
    \caption{Localization results on the RobotCar Seasons dataset. We report the recall [$\%$] at different distance and orientation thresholds, i.e., $\Delta t$ and $\Delta r$, on different conditions. i.e., dusk, sun, night, and night-rain.} %
    \label{tab:localization}
    \resizebox{0.8\textwidth}{!}{
    \begin{tabular}{*{1}{l}||*{3}{c}||*{3}{c}||*{3}{c}||*{3}{c}}
        \toprule
        \multicolumn{1}{c||}{}& \multicolumn{3}{c||}{dusk}&\multicolumn{3}{c||}{sun}&\multicolumn{3}{c||}{night}&\multicolumn{3}{c}{night-rain}\\\cline{2-13}
        \multicolumn{1}{r||}{$\Delta t$@} & \multicolumn{3}{c||}{.25 / .50 / 5.0} & \multicolumn{3}{c||}{.25 / .50 / 5.0}& \multicolumn{3}{c||}{.25 / .50 / 5.0}& \multicolumn{3}{c}{.25 / .50 / 5.0}\\
        \multicolumn{1}{r||}{Method \qquad \qquad $\Delta r$@}  & \multicolumn{3}{c||}{2 / 5 / 10} & \multicolumn{3}{c||}{2 / 5 / 10}& \multicolumn{3}{c||}{2 / 5 / 10}& \multicolumn{3}{c}{2 / 5 / 10}\\
        \midrule
        AS &  \multicolumn{3}{c||}{44.7 / 74.6 / 95.9} & \multicolumn{3}{c||}{25.0 / 46.5 / 69.1} &\multicolumn{3}{c||}{0.5 / 1.1 / 3.4} &\multicolumn{3}{c}{1.4 / 3.0 / 5.2} \\
        CSL &  \multicolumn{3}{c||}{56.6 / 82.7 / 95.9} & \multicolumn{3}{c||}{28.0 / 47.0 / 70.4} &\multicolumn{3}{c||}{0.2 / 0.9 / 5.3} &\multicolumn{3}{c}{0.9 / 4.3 / 9.1} \\
        NV+SIFT & \multicolumn{3}{c||}{55.6 / 83.5 / 95.3} & \multicolumn{3}{c||}{46.3 / 67.4 / 90.9} &\multicolumn{3}{c||}{4.1 / 9.1 / 24.4} &\multicolumn{3}{c}{2.3 / 10.2 / 20.5} \\
        NV+SP & \multicolumn{3}{c||}{54.8 / 83.0 / 96.2} & \multicolumn{3}{c||}{51.7 / 73.9 / 92.4} &\multicolumn{3}{c||}{6.6 / 17.1 / 32.2} &\multicolumn{3}{c}{5.2 / 17.0 / 26.6}  \\
        NV+SP+SG & \multicolumn{3}{c||}{62.4 / 83.5 / 97.2} & \multicolumn{3}{c||}{52.8 / 80.2 / 96.1} &\multicolumn{3}{c||}{29.0 / 66.9 / 90.2} &\multicolumn{3}{c}{46.4 / 76.1 / 92.0} \\
        HF-Net & \multicolumn{3}{c||}{53.9 / 81.5 / 94.2} & \multicolumn{3}{c||}{48.5 / 69.1 / 85.7} &\multicolumn{3}{c||}{2.7 / 6.6 / 15.8 } &\multicolumn{3}{c}{4.7 / 16.8 / 21.8}  \\
        \hline
        SuperGF-SparseSIFT& \multicolumn{3}{c||}{55.8 / 83.6 / 95.0} & \multicolumn{3}{c||}{45.8 / 67.2 / 91.1} &\multicolumn{3}{c||}{3.6 / 8.8 / 20.8} &\multicolumn{3}{c}{2.6 / 11.5 / 19.8}\\
        SuperGF-DenseSIFT & \multicolumn{3}{c||}{56.2 / 84.2 / 97.2} & \multicolumn{3}{c||}{48.4 / 68.9 / 91.2} &\multicolumn{3}{c||}{6.9 / 17.9 / 33.1} &\multicolumn{3}{c}{6.2 / 17.3 / 32.6} \\
        SuperGF-SparseSP & \multicolumn{3}{c||}{55.1 / 83.0 / 96.7} & \multicolumn{3}{c||}{52.2 / 75.7 / 96.1} &\multicolumn{3}{c||}{8.0 / 19.9 / 36.8} &\multicolumn{3}{c}{8.2 / 22.0 / 33.9} \\
        SuperGF-DenseSP & \multicolumn{3}{c||}{56.6 / 83.5 / 97.0} & \multicolumn{3}{c||}{53.1 / 76.4 / 95.8} &\multicolumn{3}{c||}{12.4 / 26.3 / 45.3} &\multicolumn{3}{c}{13.0 / 25.2 / 35.8}\\
        SuperGF-SparseSP-SG & \multicolumn{3}{c||}{64.2 / 84.3 / 98.5} & \multicolumn{3}{c||}{54.2 / 81.5 / 97.1} &\multicolumn{3}{c||}{33.6 / 69.9 / 92.6} &\multicolumn{3}{c}{49.2 / 78.1 / 92.4}\\
        SuperGF-DenseSP-SG & \multicolumn{3}{c||}{\textbf{65.1} / \textbf{84.5} / \textbf{98.5}} & \multicolumn{3}{c||}{\textbf{55.8} / \textbf{82.5} / \textbf{97.5}} &\multicolumn{3}{c||}{\textbf{34.2} / \textbf{70.3} / \textbf{93.0}} &\multicolumn{3}{c}{\textbf{49.9} / \textbf{78.2} / \textbf{92.8}}
\\\bottomrule
\end{tabular}
    }
\end{table*}

\subsection{Visual Localization}
We report the pose recall at position and orientation thresholds different for each sequence, as defined by the benchmark~\cite{robotcar}. Table~\ref{tab:localization} shows the localization results for the different methods. We can make the following observations. 

First, compared to methods that intermediate a retrieval step, structure-based methods, i.e., AS and CSL, show competitive results on the dusk sequence, where the accuracy tends to saturate. In the more challenging sequences, methods based on a hierarchical approach tend to work significantly better than structure-based methods, which suffer from the increased ambiguity of the matches. 

Next, in contrasting the two pipelines of NV+SIFT and NV+SP, we can observe that NV+SIFT shows slightly better results than NV+SP on easy cases, such as dusk. In contrast, the NV+SP shows significantly better results than NV+SIFT in those challenging cases, especially at night. It indicates that the SIFT descriptors perform better than SuperPoint on camera pose estimation in normal cases, but the conclusion is the opposite in the situation of challenging cases. It is consistent with the findings of previous studies~\cite{song2021matching,jin2020image}.

Finally, by comparing our method with baselines, we can observe that although the two-stage localization fuses image retrieval and camera pose estimation, it is clear that the latter plays a dominant role. Especially for simple cases, the use of SuperGF, which can be regarded as a robust global feature, for retrieval has limited improvement for hierarchical localization. But for challenging cases, we can observe the localization improvement using SuperGF. Overall, by implanting SuperGF, our pipelines achieve state-of-art performance on 6-DoF localization compared with others.

\begin{table}\centering
    \caption{Comparisons of model size and latency. Latency is measured on an NVIDIA TITAN RTX GPU.} %
    \label{tab:latency}
    \resizebox{1\linewidth}{!}{
    \begin{tabular}{*{1}{l}||c|c}
        \toprule
       Method& Memory (MB)&Extraction latency (ms)\\\midrule
       NetVLAD& 14.8 & 15.5\\
       SFRS& 149.0 & 16.4\\
       ResNet50-GeM & 23.5 & 11.1\\
       SOLAR & 56.2 & 19.5\\\hline
       \textbf{SuperGF-dense} & 6.1 & 4.4\\
       \textbf{SuperGF-sparse} & 2.2 & 2.3
\\\bottomrule
\end{tabular}
    }
\end{table}
\subsection{Latency and Memory}

In real-world applications, latency and scalability are important factors that need to be considered for real-time visual localization. We report the computational time and memory requirements for compared methods of extracting glocal image features to process a single query image; see Table~\ref{tab:latency}. In practice, we unify the input image size as $480 \times 640$ for all methods. In the case of SuperGF-sparse, we randomly selected 100 images to generate sparse local features for the test.  SuperGF shows clear advantages in both terms than others. 

Combining the results of previous sections, we can conclude that SuerGF uses minimal resources but generates SOTA-level global image features for image retrieve and lead to better performance of 6-DoF localization. In particular, the sparse version of SuperGF causes a loss of retrieval accuracy to some extent, but it has obvious advantages in terms of feature extraction efficiency. It is worth noting that SuperGF is affected by the adopted local descriptor significantly. We prefer to use learning-based local features, such as SuperPoint, joined with SuperGF. We tried to use SuperGF to generate global features for hand-craft local features, i.e., SIFT, but the results were unsatisfactory.  One possible reason is the limited representative ability of the hand-craft local features, which only jointly low-level information of images. 

\section{Summary and Conclusion}
This paper has presented a novel method for global feature extraction, i.e., SuperGF. It is transformer-based and designed for 6-DoF localization, which acts directly on local features provided by descriptors of image-matching. The results show our method's advantages in terms of accuracy and efficiency. 

We provide different implementations of SuperGF. i.e., different types of local features with both learning-based and hand-craft descriptors. Users can choose different versions according to their needs. Moreover, we encourage using the learning-based descriptor, i.e., SuperPoint, joint with SuperGF. In addition, if you pursue higher retrieval accuracy while localization, e.g., using for loop closure, then the dense version is more suitable. For 6-DoF localization alone, the sparse version is better in efficiency.


{\small
\bibliographystyle{ieee_fullname}
\bibliography{egbib}
}

\end{document}